\documentclass[runningheads]{paper301}

\usepackage{multirow}
\usepackage{amsmath,graphicx,booktabs,amssymb,stmaryrd}
\usepackage{caption}
\usepackage{color}
\usepackage{bbm}
\usepackage{booktabs}
\usepackage{bbding}
\usepackage[pagebackref=true,breaklinks=false,colorlinks,bookmarks=false]{hyperref}

\begin{document}

\title{OXnet: Deep Omni-supervised Thoracic Disease Detection from Chest X-rays}

\author{Luyang Luo*\inst{1}\Envelope, Hao Chen*\inst{2},\\
Yanning Zhou\inst{1}, Huangjing Lin\inst{1,3}, \and Pheng-Ann Heng\inst{1,4}}
% index{Luo, Luyang}
% index{Chen, Hao}
% index{Zhou, Yanning}
% index{Lin, Huangjing}
% index{Heng, Pheng-Ann}

\institute{$^1$Department of Computer Science and Engineering,\\
The Chinese University of Hong Kong, Hong Kong, China\\
\email{lyluo@cse.cuhk.edu.hk}\\
$^2$Department of Computer Science and Engineering,\\
The Hong Kong University of Science and Technology, Hong Kong, China.\\
\email{jhc@cse.ust.hk}\\
$^3$Imsight AI Research Lab, Shenzhen, China\\
$^4$Guangdong-Hong Kong-Macao Joint Laboratory of Human-Machine Intelligence-Synergy Systems, Shenzhen Institutes of Advanced Technology,\\ Chinese Academy of Sciences, China}

\authorrunning{Luyang Luo, Hao Chen, et al.}

\titlerunning{OXnet: Omni-supervised Thoracic Disease Detection from Chest X-rays}

\maketitle              

\footnotetext[1]{The first two authors contributed equally.}

\begin{abstract}

Chest X-ray (CXR) is the most typical diagnostic X-ray examination for screening various thoracic diseases. 
Automatically localizing lesions from CXR is promising for alleviating radiologists' reading burden. 
However, CXR datasets are often with massive image-level annotations and scarce lesion-level annotations, and more often, without annotations.
Thus far, unifying different supervision granularities to develop thoracic disease detection algorithms has not been comprehensively addressed.
In this paper, we present OXnet, the first deep omni-supervised thoracic disease detection network to our best knowledge that \emph{uses as much available supervision as possible} for CXR diagnosis. 
We first introduce supervised learning via a one-stage detection model.
Then, we inject a global classification head to the detection model and propose dual attention alignment to guide the global gradient to the local detection branch, which enables learning lesion detection from image-level annotations. 
We also impose intra-class compactness and inter-class separability with global prototype alignment to further enhance the global information learning. 
Moreover, we leverage a soft focal loss to distill the soft pseudo-labels of unlabeled data generated by a teacher model.
Extensive experiments on a large-scale chest X-ray dataset show the proposed OXnet outperforms competitive methods with significant margins.
Further, we investigate omni-supervision under various annotation granularities and corroborate OXnet is a promising choice to mitigate the plight of annotation shortage for medical image diagnosis.\footnote[2]{Code is available at \url{https://github.com/LLYXC/OXnet}.}

\end{abstract}

\section{Introduction}
\label{sec:introduction}

Modern object detection algorithms often require a large amount of supervision signals. However, annotating abundant medical images for disease detection is infeasible due to the high dependence of expert knowledge and expense of human labor. Consequently, many medical datasets are weakly labeled or, more frequently, unlabeled \cite{tajbakhsh2020embracing}. This situation especially exists for chest X-rays (CXR), which is the most commonly performed diagnostic X-ray examination. Apart from massive unlabeled data, CXR datasets often have image-level annotations that can be easily obtained by text mining from numerous radiological reports \cite{wang2017chestx,irvin2019chexpert}, while lesion-level annotations (e.g., bounding boxes) scarcely exist \cite{huang2020rectifying,wang2021knowledge}. Therefore, efficiently leveraging available annotations to develop thoracic disease detection algorithms has significant practical value. 

Omni-supervised learning \cite{radosavovic2018data} aims to leverage the existing fully-annotated data and other available data (e.g. unlabeled data) as much as possible, which could practically address the mentioned challenge. Distinguished from previous studies that only include extra unlabeled data \cite{radosavovic2018data,huang2018omni,venturini2020uncertainty}, we target on utilizing \emph{as much available supervision as possible} from data of various annotation granularities, i.e., fully-labeled data, weakly-labeled data, and unlabeled data, to develop a \emph{unified framework} for thoracic disease detection on chest X-rays. 

In this paper, we present OXnet, a unified deep framework for omni-supervised chest X-ray disease detection. To this end, we first leverage limited bounding box annotations to train a base detector. To enable learning from weakly labeled data, we introduce a dual attention alignment module that guides gradient from an image-level classification branch to the local lesion detection branch. To further enhance learning from global information, we propose a global prototype alignment module to impose intra-class compactness and inter-class separability. 
For unlabeled data, we present a soft focal loss to distill the soft pseudo-labels generated by a teacher model. 
Extensive experiments show that OXnet not only outperforms the baseline detector with a large margin but also achieves better thoracic disease detection performance than other competitive methods. Further, OXnet also show comparable performance to fully-supervised method with fewer fully-labeled data and sufficient weakly-labeled data. In summary, OXnet can effectively utilize all the available supervision signals, demonstrating a promisingly feasible and general solution to real-world applications.

\section{Method}
\label{sec:method}
Let $\mathcal{D}_{F}$, $\mathcal{D}_{W}$, and $\mathcal{D}_{U}$ denote the fully-labeled data (with bounding box annotations), weakly-labeled data (with image-level annotations), and unlabeled data, respectively. As illustrated in Fig. \ref{Framework}, OXnet correspondingly consists of three main parts: a base detector to learn from $\mathcal{D}_{F}$, a global path to learn from $\mathcal{D}_{W}$, and an unsupervised branch to learn from $\mathcal{D}_{U}$. As the first part is a RetinaNet \cite{lin2017focal} backbone supervised by the focal loss and bounding box regression loss, we will focus on introducing the latter two parts in the following sections.

\begin{figure}[t]
\centering
\includegraphics[width = \linewidth]{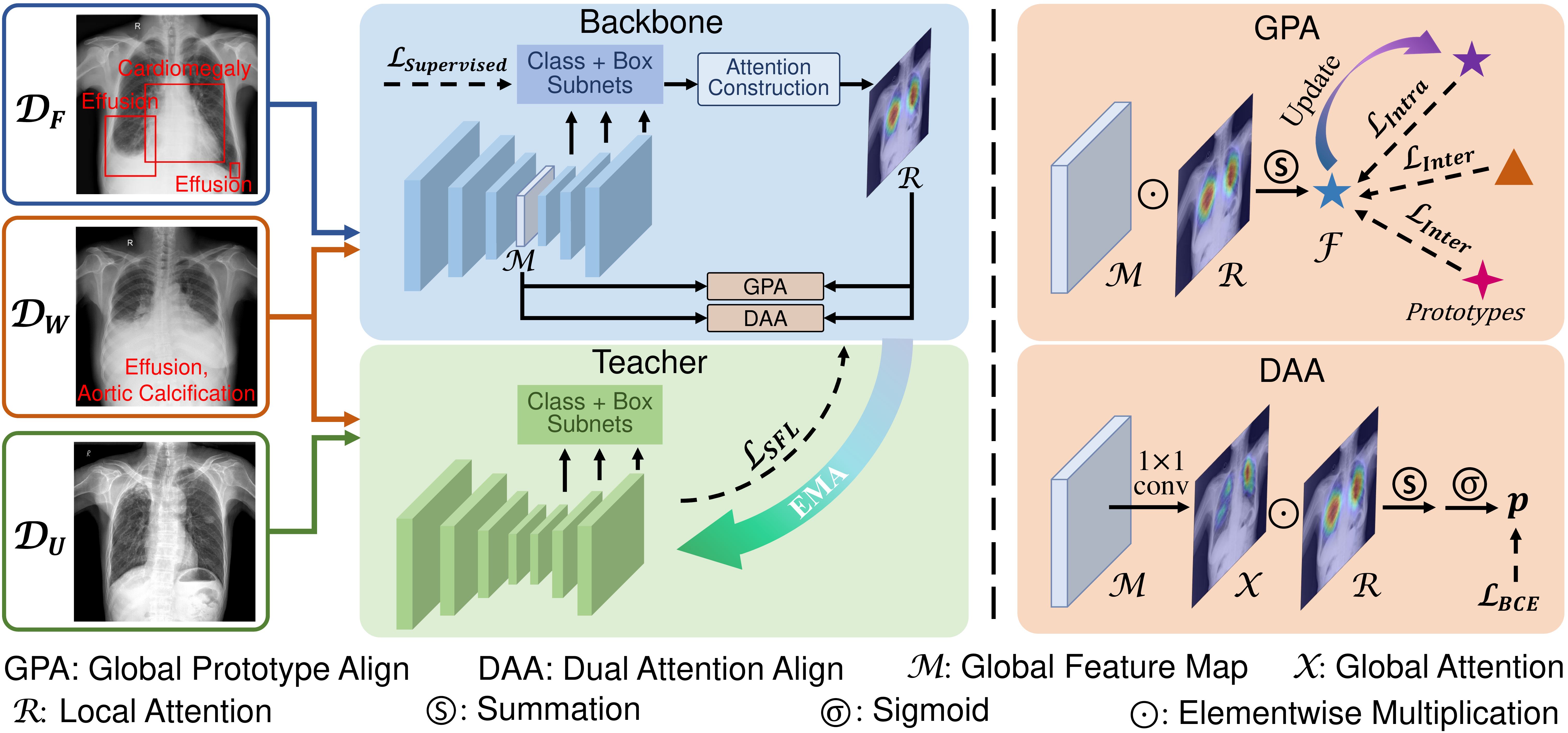}
\centering
\caption{Overview of OXnet. For $\mathcal{D}_{F}$, the model learns via the original supervised loss of RetinaNet. For $\mathcal{D}_{W}$ and $\mathcal{D}_{F}$, DAA aligns the global and local attentions and enables global information flowing to the local branch (Sec. \ref{sec:DAA}). Further, GPA enhances the learning from global labels (Sec. \ref{sec:GPA}). For learning $\mathcal{D}_{U}$ and $\mathcal{D}_{W}$, a teacher model would guide the RetinaNet with soft focal loss (Sec. \ref{sec:SFL}).}
\label{Framework}
\end{figure}

\subsection{Dual Attention Alignment for Weakly-supervised Learning}
\label{sec:DAA}

To enable learning from $\mathcal{D}_{W}$, we first add a multi-label classification global head to the ResNet \cite{he2016deep} part of RetinaNet. This global head consists of a 1$\times$1 convolutional layer for channel reduction, a global average pooling layer, and a sigmoid layer. Then, the binary cross entropy loss is used as the objective:
\begin{equation}
    \mathcal{L}_{\rm BCE}(p,y) = -\sum_{c=1}^{N} [ y_{c}\cdot {\rm log}(p_{c}) + (1-y_{c})\cdot {\rm log}(1-p_{c}) ]
\end{equation}
where $c$ is the index of total $N$ categories, $y_{c}$ is the image-level label, and $p_{c}$ is the prediction. The 1$\times$1 convolution gets input $\mathcal{M}$ and outputs feature map $\mathcal{X}$ with $N$ channels, where $\mathcal{X}$ is exactly the global attention map, i.e., class activating map \cite{zhou2016learning}. Meanwhile, the local classification head of RetinaNet conducts dense prediction and generates outputs of size Width$\times$Height$\times$Classes$\times$Anchors for each feature pyramid level \cite{lin2017feature}. These classification maps can be directly used as the local attentions for each class and anchor. Hence, we first take the maximum on the anchor dimension to get the attentions $\mathcal{A}$ on number of $M$ pyramid levels. Then, the final local attention for each class is obtained by:
\begin{equation}
    \mathcal{A}_{c} = {\rm argmax}_{\mathcal{A}_{c}}({{\rm max}(\mathcal{A}_{c}^{1}), {\rm max}(\mathcal{A}_{c}^{2}), \cdots, {\rm max}(\mathcal{A}_{c}^{M})})
\end{equation}

\begin{figure}[t]
\centering
\includegraphics[width = 0.93\linewidth]{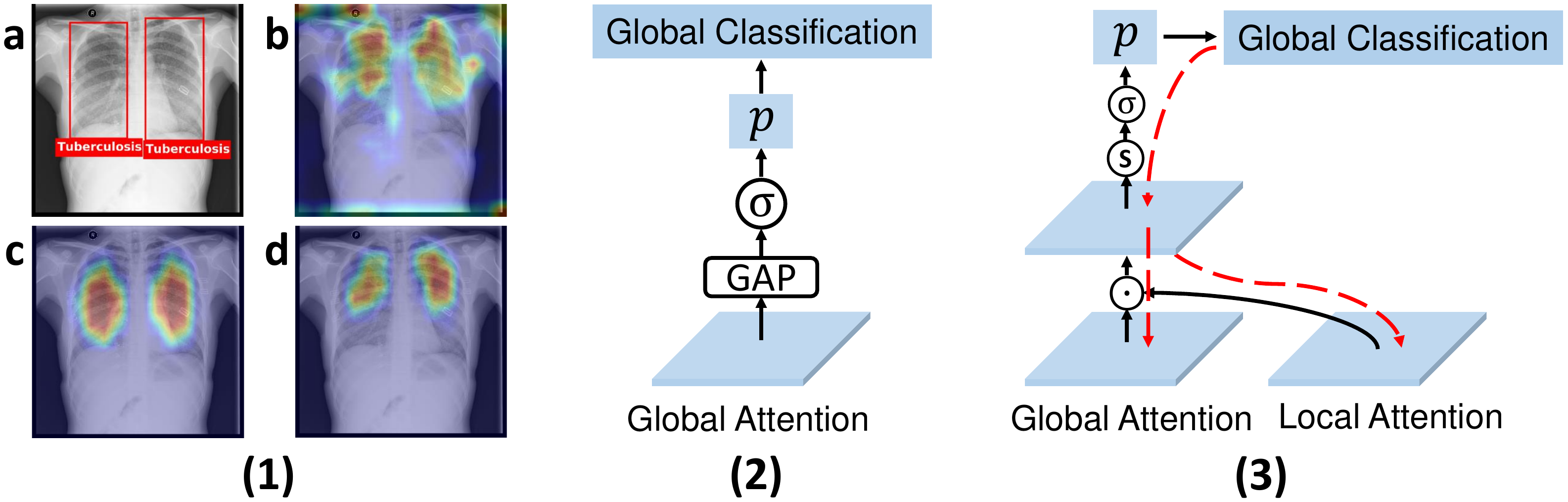}
\centering
\caption{\textbf{(1)}. (a) The ground truth of lesions; (b) The global attention without DAA; (c) The local attention map; (d) The global attention with DAA. \textbf{(2)}. The global classification head without DAA. \textbf{(3)}. The DAA module, where solid line and \textcolor{red}{dashed line} represent forward and backward flow, respectively.}
\label{Attention_Mismatch}
\end{figure}

Essentially, we observed that the global and local attentions are often mismatched as shown in Fig. \ref{Attention_Mismatch} (1), indicating the two branches making decisions out of different regions of interests (ROIs). Particularly, the local attention usually covers more precise ROIs by learning from lesion-level annotations. Therefore, we argue that the local attention can be used to weigh the importance of each pixel on $\mathcal{X}_{c}$.
We then resize the local attention $\mathcal{A}_{c}$ to be the same shape as $\mathcal{X}_{c}$ and propose a dual attention alignment (DAA) module as follows:
\begin{equation}
    p_{c} = \sigma \left (\sum_{i} [ \mathcal{X}_{c} \odot \mathcal{R}_{c} ]_{i}\right ) \text{ ,     } \mathcal{R}_{c} = \frac{\mathcal{A}_{c}}{\sum_{i}[\mathcal{A}_{c}]_{i}}
\end{equation}
where $\sigma$ is the sigmoid function, $i$ is the index of the pixels, and $\odot$ denotes element-wise multiplication. Particularly, we let $\sum_{i}\mathcal{R}_{c} = 1$, so the element-wise multiplication constructs a pooling function as in the multiple instance learning literature \cite{ilse2018attention,tang2017multiple}. Then, DAA is used to replace the mentioned global head as illustrated in Fig. \ref{Attention_Mismatch} (2) and (3). Consequently, the local attention helps rectify the decision-making regions of the global branch, and the gradient from the global branch could thus flow to the local branch. To this stage, the local classification branch not only receives the strong yet limited supervision from $\mathcal{D}_{F}$ but also massive weak supervision from $\mathcal{D}_{W}$. 

\subsection{Global Prototype Alignment for Multi-label Metric Learning}
\label{sec:GPA}
With DAA, any global supervision signals could flow to the local branch. To further enhance this learning process, we propose a multi-label metric learning loss that imposes intra-class compactness and inter-class separability to the network. It is worth noting that multi-label metric learning can be hard as the deep features often capture multi-class information \cite{luo2020deep}. Specially, in our case, we can obtain \emph{category-specific} features under the guidance of local attention as follows:

\begin{equation}
    \mathcal{F}_{c} = \sum_{i}\mathcal{M}\odot \mathcal{R}_{c}
\end{equation}
where $\mathcal{M}$ is the feature map before the global 1$\times$1 convolutional layer. Particularly, $\mathcal{R}_{c}$ is of shape $W\times H$ with $W$ and $H$ being width and height, respectively, $i$ is the index of the feature vectors ($\mathcal{M}$ contains \#$W\times H$ vectors), and $\odot$ is the element-wise multiplication over the dimensions of $W$ and $H$. Here, the local attention $\mathcal{R}_{c}$ highlights the feature vectors related to each category, leading to the aggregated category-specific feature vector $\mathcal{F}_{c}$. For better metric learning regularization, we generate global prototypes \cite{wen2016discriminative,yang2018robust} for each category as follows: 

\begin{equation}
    \mathcal{P}_{t+1} = \beta\cdot \mathcal{P}_{t} + (1-\beta) \cdot \frac{\sum_{k}^{K}p_{k}\cdot\mathcal{F}_{k}}{\sum_{k}^{K}p_{k}}
\end{equation}

\noindent where $\beta$ is 0 if $t=0$ otherwise 0.7, and $K$ is the number of data in a batch. Particularly, the prototype is updated with the weighted average of confidences \cite{shi2020towards,xu2020cross}. We then align $\mathcal{F}_{c}$ with the prototypes as follows:
\begin{equation}
\begin{gathered}
    \mathcal{L}_{\rm Intra} = \frac{1}{N}\sum_{c=1}^{N}\|\mathcal{F}_{c}- \mathcal{P}_{c} \|_{2}^{2} \\
    \mathcal{L}_{\rm Inter} = \frac{1}{N(N-1)}\sum_{c=1}^{N} \sum_{0\leq j\neq c\leq N}^{N} {\rm max}(0, \delta-\|\mathcal{F}_{j}- \mathcal{P}_{c} \|_{2}^{2})
\end{gathered}
\end{equation}
where $\delta$ is a scalar representing the inter-class margin.

\subsection{Soft Focal Loss for Unsupervised Knowledge Distillation}

\label{sec:SFL}
To learn from the unlabeled data, we first obtain a mean teacher model by exponential moving average \cite{tarvainen2017mean}: $\tilde{\theta}_{t+1} = \lambda\cdot \tilde{\theta}_{t} + (1-\lambda)\cdot \theta_{t+1}$, where $\theta$ and $\tilde{\theta}$ are the parameters of RetinaNet and the teacher model, respectively, $\lambda$ is a decay coefficient, and $t$ represents training step. Then, the student RetinaNet will learn from the teacher model's classification predictions, which are probabilities in range (0, 1).
Particularly, the foreground-background imbalance problem \cite{lin2017focal} should also be taken care of.
Therefore, we extend the focal loss to its soft form inspired by \cite{li2020generalized,wang2020focalmix}. Specifically, the original focal loss is as follows:
\begin{equation}
    \mathcal{L}_{\rm FL}(q) = - \alpha\cdot (1-q)^{\gamma}\cdot log(q)
\label{focal_loss}
\end{equation}
where $\alpha$ is a pre-set weight for balancing foreground and background, $q$ is the local classification prediction from the model (we eliminate the subscript $c$ for simplicity), $\gamma$ is a scalar controlling $(1-q)^{\gamma}$ to assign more weights onto samples which are less-well classified. 
We obtain the soft focal loss as follows:
\begin{equation}
    \mathcal{L}_{\rm SFL}(q,\tilde{q}) = -(\tilde{q}\alpha + \epsilon)\cdot\|\tilde{q}-q\|^{\gamma}\cdot \mathcal{L}_{\rm BCE}(q,\tilde{q})
\end{equation}

Here, instead of assigning $\alpha$ according to whether an anchor is positive or negative, we assume its value changes linearly with the value of teacher model's prediction $\tilde{q}$ and modify it to be $\tilde{q}\alpha + \epsilon$, where $\epsilon$ is a constant. Meanwhile, we notice that $(1-q)^{\gamma}$ in Eq. (\ref{focal_loss}) depends on how closer $q$ is to its target ground truth. Therefore, we modify the focal weight to be $\|\tilde{q}-q\|^{\gamma}$ to give larger wights to the instances with higher disagreement between student-teacher models.

\subsection{Unified Training for Deep Omni-supervised Detection}

The overall training loss sums up the supervised losses \cite{lin2017focal}, weakly-supervised losses ($\mathcal{L}_{\rm BCE}$, $\mathcal{L}_{\rm Intra}$, and $\mathcal{L}_{\rm Inter}$), and unsupervised loss ($\mathcal{L}_{\rm SFL}$). Particularly, weakly-supervised losses are applied to $\mathcal{D}_{W}$ and $\mathcal{D}_{F}$, and unsupervised loss is applied to $\mathcal{D}_{U}$ and $\mathcal{D}_{W}$. 
Note that there were works on the NIH dataset \cite{wang2017chestx} leveraging two types of annotations \cite{li2018thoracic,liu2019align,ouyang2020learning,zhou2021contrast} for lesion localization. However, these methods localize lesions with attention maps, which cannot be easily compared to bounding box-based detection model that we aim to develop.

$\beta$ is set to 0 for the first step otherwise 0.7, $\delta$ is set to 1, $\lambda$ is set to 0.99 by tuning on the validation set. As suggested in \cite{wang2020focalmix}$, \alpha$ is set to 0.9,  $\epsilon$ is 0.05, and $\gamma$ is set to 2. CXR images are resized to 512$\times$512 without cropping. Data augmentation is done following \cite{zhou2020deep}. Adam \cite{kingma2015adam} is used as the optimizer. The learning rate is initially set to 1e-5 and divided by ten when validation mAP stagnate. All implementations use Pytorch \cite{NEURIPS2019_pytorch} on an NVIDIA TITAN Xp GPU.

\section{Experiments}
\label{sec:experiments}

\subsection{Dataset and Evaluation Metrics}
In total, 32,261 frontal CXR images taken from 27,253 patients were used. 
A cohort of 10 radiologists (4-30 years of experience) was involved in labeling the lesion bounding boxes, and each image was labeled by two with corresponding text report. 
If the initial labelers disagreed on the annotation, a senior radiologist ($\geq$ 20-year experience) would make the final decision.
Nine thoracic diseases including aortic calcification, cardiomegaly, fracture, mass, nodule, pleural effusion, pneumonia, pneumothorax, and tuberculosis were finally chosen in this study by the consensus of the radiologists.
The dataset was split into training, validation, and testing sets with 13,963, 5,118, and 13,180 images, respectively, without patients overlapping. Detailed numbers of images and annotations per set can be found in the supplementary.

We used Average Precision (AP) \cite{everingham2010pascal} as the evaluation metric. Specifically, we reported the mean AP (mAP) from AP$^{40}$ to AP$^{75}$ with an interval of 5, following the Kaggle Pneumonia Detection Challenge\footnote{\url{https://www.kaggle.com/c/rsna-pneumonia-detection-challenge}} and \cite{gabruseva2020deep}. We also evaluated AP$^{\rm S}$, AP$^{\rm M}$, and AP$^{L}$ for small, medium, and large targets, respectively, using the COCO API\footnote{\url{https://cocodataset.org/\#detection-eval}}. All reported statistics were averaged over the nine abnormalities.

\subsection{Comparison with Other Methods}

To our best knowledge, few previous works simultaneously leveraged fully-labeled data, weakly-labeled data, and unlabeled data. Hence, we first trained a RetinaNet (with pre-trained Res101 weight from ImageNet \cite{deng2009imagenet}; a comparison of baselines could be found in the supplementary) on our dataset. Then, we implemented several state-of-the-art semi-supervised methods finetuned from RetinaNet, including RetinaNet \cite{lin2017focal}, $\rm \Pi$ Model \cite{laine2016temporal}, Mean Teacher \cite{tarvainen2017mean}, MMT-PSM \cite{zhou2020deep}, and FocalMix \cite{wang2020focalmix}. We further enabled learning from the weak annotations by adding a global classification head to each semi-supervised method to construct multi-task (MT) models.

\begin{table}[t!]
\caption{Quantitative comparison of different methods.}
\centering
\resizebox{\textwidth}{!}{
\begin{tabular}{c|c|c|c|cccccc}
\hline
\hline
\multirow{3}{*}{\textbf{Method}} & \multicolumn{3}{c|}{\textbf{\# images used}}                                                                                                                                         & \multicolumn{6}{c}{\textbf{Metrics}}            \\ \cline{2-10} 
                        & \multicolumn{1}{c|}{\begin{tabular}[c]{@{}c@{}}Fully \\ Labeled\end{tabular}} & \multicolumn{1}{c|}{\begin{tabular}[c]{@{}c@{}}Weakly \\ Labeled\end{tabular}} & Unlabeled & mAP   & AP$^{40}$ & AP$^{75}$ & AP$^{\rm S}$  & AP$^{\rm M}$  & AP$^{\rm L}$  \\ \hline
RetinaNet \cite{lin2017focal}               & 2725                                                    & 0                                                        & 0         & 18.4 & 27.7 & 7.5  & 8.0    & 16.7 & 25.4 \\ \hline
$\rm \Pi$ Model \cite{laine2016temporal}                & 2725                                                    & 0                                                        & 11238     & 20.0 & 29.3 & 9.3  & 9.2  & 20.8 & 27.0 \\
Mean Teacher \cite{tarvainen2017mean}            & 2725                                                    & 0                                                        & 11238     & 20.0 & 29.2 & 9.4  & 9.1  & 20.4 & 26.9 \\
MMT-PSM \cite{zhou2020deep}                 & 2725                                                    & 0                                                        & 11238     & 19.1     & 28.4     & 8.4     & 8.8     & 19.3     & 26.5     \\
FocalMix \cite{wang2020focalmix}                & 2725                                                    & 0                                                        & 11238     & 19.8 & 29.1 & 9.0  & 8.6  & 19.6 & 26.3 \\ \hline
$\rm \Pi$ Model \cite{laine2016temporal} + MT           & 2725                                                    & 11238                                                    & 0         & 20.2 & 29.6 & 9.2  & 9.4  & 21.3 & 27.6 \\
Mean Teacher \cite{tarvainen2017mean} + MT       & 2725                                                    & 11238                                                    & 0         & 20.4 & 29.6 & 9.4  & 9.3  & 20.6 & 27.1 \\
MMT-PSM \cite{zhou2020deep} + MT            & 2725                                                    & 11238                                                    & 0         & 19.3     & 28.4     & 8.8     & 8.4     & 17.9     & 26.8     \\
FocalMix \cite{wang2020focalmix} + MT           & 2725                                                    & 11238                                                    & 0         & 20.1 & 29.7 & 8.8  & 8.5  & 18.3 & 27.4 \\ \hline
SFL                     & 2725                                                    & 0                                                        & 11238     & 20.4 & 29.7 & 9.4  & 9.3  & 21.6 & 29.8 \\
DAA                & 2725                                                    & 11238                                                    & 0         & 21.2 & 31.2 & 9.4  & 9.8 & 20.7 & 28.0 \\
DAA + GPA                & 2725                                                    & 11238                                                    & 0         & 21.4 & 31.4 & 9.5  & \textbf{11.0} & 20.9 & 27.1 \\
OXnet (SFL+DAA+GPA)         & 2725                                                    & 11238                                                    & 0         & \textbf{22.3} & \textbf{32.4} & \textbf{10.3} & 9.6  & \textbf{21.8} & \textbf{31.4} \\ \hline\hline
\end{tabular}
\label{tab:comparison&ablation}
}
\end{table}

\subsubsection{Quantitative Results. }All semi-supervised methods (row 2 to 5 in Table \ref{tab:comparison&ablation}) clearly outperform the RetinaNet baseline (1st row), demonstrating effective utilization of the unlabeled data. After incorporating the global classification heads, the four multi-task networks (row 6 to 9) get further improvement with about 0.2$\sim$0.4 points raising in mAP. This finding suggests that large image-level supervision can help learning abnormalities detection, but the benefits are still limited without proper design. On the other hand, OXnet achieves 22.3 in mAP and outperforms the multi-task models on various sizes of targets. The results corroborate that our proposed method can more effectively leverage the less well-labeled data for the thoracic disease detection task.

\begin{figure}[t!]
\centering
\includegraphics[width = \linewidth]{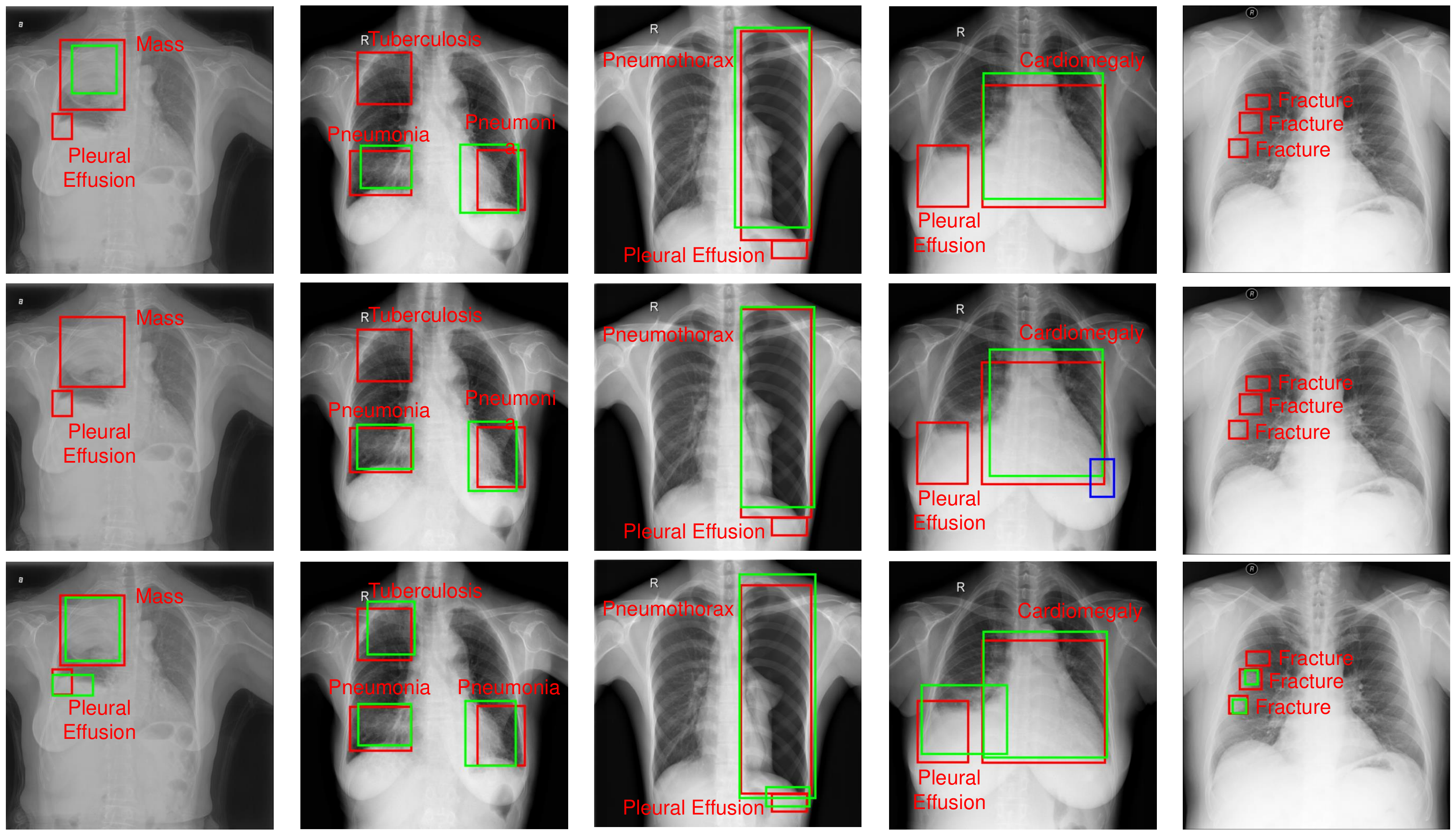}
\centering
\caption{Visualization of results generated by RetinaNet (first row), Mean Teacher + MT (second row), and our method (third row). \textcolor{red}{Ground truth is in red}, \textcolor{green}{true positives are in green}, and \textcolor{blue}{false positives are in blue}. Best viewed in color.}
\label{Qualitative}
\end{figure}

\subsubsection{Ablation Study. }The last four rows in Table \ref{tab:comparison&ablation} report the ablation study of the proposed components, i.e., soft focal loss (SFL), dual attention alignment (DAA), and global prototype alignment (GPA). Our SFL achieves an mAP of 20.4 and outperforms other semi-supervised methods, demonstrating its better capability of utilizing the unlabeled data. On the other hand, incorporating only DAA reaches an mAP of 21.2, showing effective guidance from the weakly labeled data. Adding GPA to DAA improves the mAP to 21.4, demonstrating the effectiveness of the learned intra-class compactness and inter-class separability. By unifying the three components, OXnet reaches the best results in 5 out of 6 metrics, corroborating the complementarity of the proposed methods.

\subsubsection{Qualitative Results. }We also visualize the outputs generated by RetinaNet, Mean Teacher + MT (the best method among those compared with ours), and OXnet in Fig. \ref{Qualitative}. As illustrated, our model yields more accurate predictions for multiple lesions of different diseases in each chest X-ray sample. We also illustrates more samples of the attention maps in the supplementary.

\begin{table}[b!]
\caption{Results under different ratios of annotation granularities.}
\centering
\begin{tabular}{c|c|c|c|cccccc}
\hline\hline
\multirow{3}{*}{\textbf{Method}}    & \multicolumn{3}{c|}{\textbf{\# images used}}                                                                                   & \multicolumn{6}{c}{\textbf{Metrics}}                                     \\ \cline{2-10} 
                           & \begin{tabular}[c]{@{}c@{}}Fully\\ Labeled\end{tabular} & \begin{tabular}[c]{@{}c@{}}Weakly\\ Labeled\end{tabular} & Unlabeled & mAP  & AP$^{40}$ & AP$^{75}$ & AP$^{\rm S}$ & AP$^{\rm M}$ & AP$^{\rm L}$ \\ \hline
\multirow{3}{*}{RetinaNet} & 682                                                     & 0                                                        & 0         & 12.5 & 18.9      & 5.5       & 7.0          & 16.4         & 18.1         \\
                           & 1372                                                    & 0                                                        & 0         & 14.6 & 21.8      & 6.5       & 5.8          & 10.4         & 19.5         \\
                           & 2725                                                    & 0                                                        & 0         & 18.4 & 27.7      & 7.5       & 8.0          & 16.7         & 25.4         \\ \hline
\multirow{6}{*}{\textbf{OXnet (Ours)}}     & 682                                                     & 13281                                                    & 0         & 14.9 & 22.6      & 6.6       & 8.4          & 16.8         & 20.3         \\
                           & 1372                                                    & 12591                                                    & 0         & 17.8 & 26.9      & 8.0       & 6.8          & 16.2         & 24.4         \\
                           & 2725                                                    & 11238                                                    & 0         & 22.3 & 32.4      & 10.3      & 9.6          & 21.8         & 31.4         \\ 
                           & 2725                                                    & 8505                                                     & 2733      & 21.9 & 32.0      & 10.0      & 9.7          & 21.6         & 29.9         \\

                           & 2725                                                    & 2733                                                     & 8505      & 21.2 & 31.0      & 10.0      & 9.0          & 22.2         & 29.0         \\
                           & 2725                                                    & 0                                                        & 11238     & 20.7 & 30.3      & 9.4       & 8.3          & 20.7         & 28.3         \\
\hline\hline
\end{tabular}
\label{tab:omni_supertvision}
\end{table}

\subsection{Omni-supervision under Different Annotation Granularities}
Efficient learning is crucial for medical images as annotations are extremely valuable and scarce. Thus, we investigate the effectiveness of OXnet given different annotation granularities. With results in Table \ref{tab:omni_supertvision}, we find that: (1) Finer annotations always lead to better performance, and OXnet achieves consistent improvements as the annotation granularity becomes finer (row 1 to 9); (2) Increasing fully labeled data benefits OXnet more (mAP improvements are 2.9 from row 4 to 5 and 4.5 from row 5 to 6) than RetinaNet (mAP improvements are 2.1 from row 1 to 2 and 3.8 from row 2 to 3); and (3) With less fully-labeled data and more weakly-labeled data, OXnet can achieve comparable performance to RetinaNet (row 2 vs. row 4, row 3 vs. row 5). These findings clearly corroborate OXnet's effectiveness in utilizing as much available supervision as possible. Moreover, unlabeled data are easy to acquire without labeling burden, and weakly-labeled data can also be efficiently obtained with natural language processing methods \cite{wang2017chestx,irvin2019chexpert,bustos2020padchest}. Therefore, we believe OXnet could serve as a promisingly feasible and general approach to real-world clinic applications.

\section{Conclusion}
\label{sec:conclusion}

We present OXnet, a deep omni-supervised learning approach for thoracic disease detection from chest X-rays. The OXnet simultaneously utilizes well-annotated, weakly-annotated, and unlabeled data as a unified framework. Extensive experiments have been conducted and our OXnet has demonstrated superiority in effectively utilizing various granularities of annotations. In summary, the proposed OXnet has shown as a promisingly feasible and general solution to real-world applications by leveraging as much available supervision as possible.

\subsubsection{Acknowledgement.} This work was supported by Key-Area Research and Development Program of Guangdong Province, China (2020B010165004), Hong Kong Innovation and Technology Fund (Project No. ITS/311/18FP and Project No. ITS/426/17FP.), and National Natural Science Foundation of China with Project No. U1813204.

\bibliographystyle{paper301}
\bibliography{paper301}

\newpage

\section{Supplementary Materials}

\begin{figure}[t]
\centering
\includegraphics[width = \linewidth]{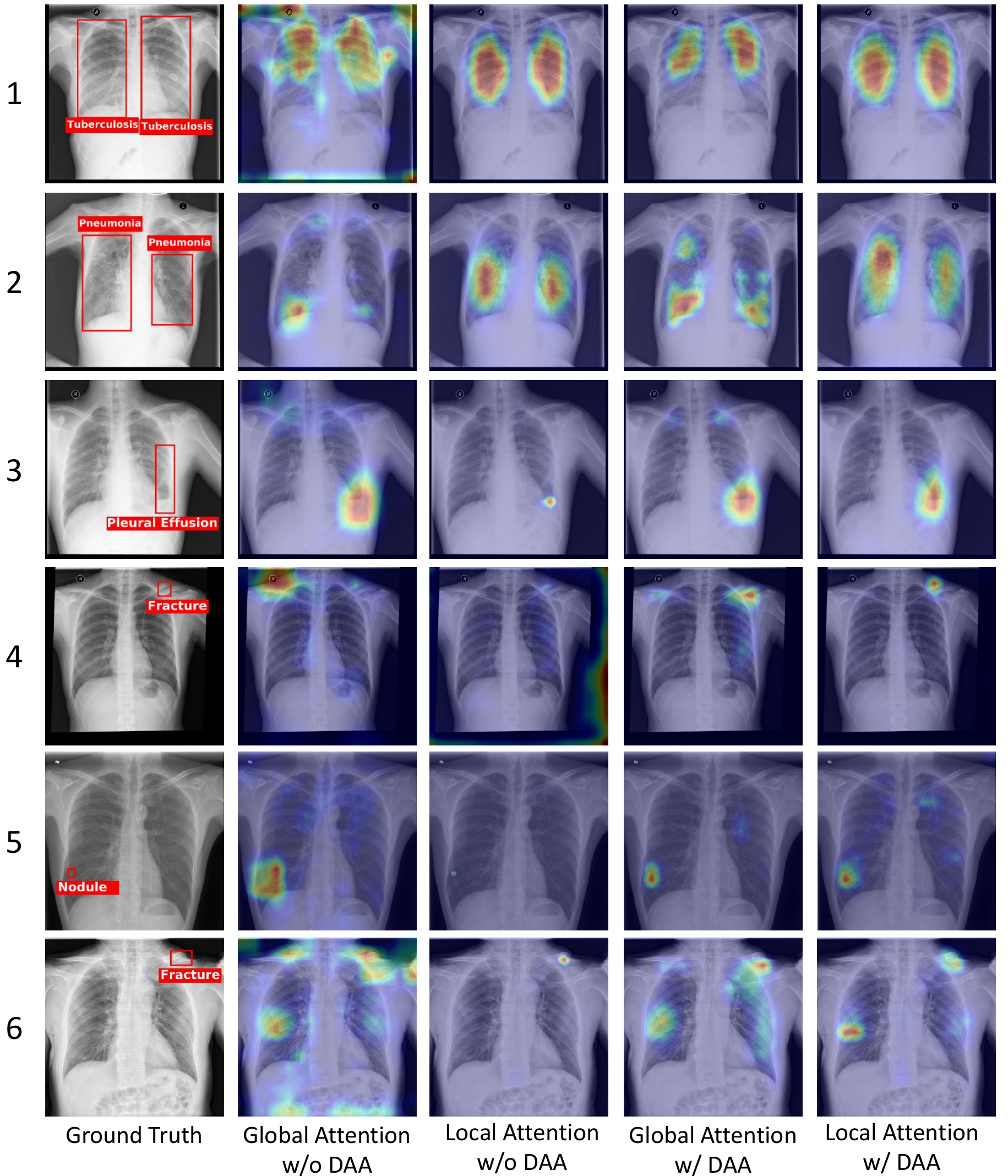}
\centering
\caption{Samples of the global attentions and local attentions. Qualitatively, it can be seen that: (a) local attention often helps refine CAM \cite{zhou2016learning} (row 1 and 2), but (b) sometimes CAM covers more accurate lesion regions (row 3); (c) Joint learning by DAA could refine both attentions (row 4); and (d) CAM covers unnecessary larger regions for very small lesions, and DAA could lead to an averaged result of both attentions (row 5 and 6).}
\label{Framework}
\end{figure}

\begin{table}[]
\centering
\setlength\tabcolsep{6pt}
\caption{Splitting of images and bounding-box annotations.}
\footnotesize
\begin{tabular}{@{}l|rrr|rrr@{}}
\toprule
                    & \multicolumn{3}{c|}{Images} & \multicolumn{3}{c}{Annotations} \\ \midrule
Pathology           & Train    & Val     & Test   & Train      & Val      & Test     \\\hline
AorticCalcification & 900      & 341     & 825    & 939        & 348      & 846      \\
Cardiomegaly        & 1098     & 388     & 1003   & 1098       & 388      & 1003     \\
Fracture            & 710      & 282     & 617    & 1893       & 707      & 1635     \\
Pleural Effusion    & 2344     & 855     & 1985   & 2899       & 1064     & 2500     \\
Mass                & 479      & 179     & 487    & 531        & 196      & 532      \\
Nodule              & 1832     & 696     & 1711   & 2777       & 1131     & 2604     \\
Pneumonia           & 2438     & 932     & 2153   & 3477       & 1307     & 3111     \\
Pneumothorax        & 1377     & 422     & 1030   & 1508       & 478      & 1154     \\
Tuberculosis        & 4455     & 1550    & 3899   & 7078       & 2525     & 6174     \\ \bottomrule
\end{tabular}
\label{tab:dataset}
\end{table}

\end{document}